# Sensor Validation using Dynamic Belief Networks

A. E. Nicholson and J. M. Brady
Engineering Science Department,
Oxford University
Parks Road, Oxford,
OX1 3PJ, UK

## Abstract

The trajectory of a robot is monitored in a restricted dynamic environment using light beam sensor data. We have a Dynamic Belief Network (DBN), based on a discrete model of the domain, which provides discrete monitoring analogous to conventional quantitative filter techniques. Sensor observations are added to the basic DBN in the form of specific evidence. However, sensor data is often partially or totally incorrect. We show how the basic DBN, which infers only an impossible combination of evidence, may be modified to handle specific types of incorrect data which may occur in the domain. We then present an extension to the DBN, the addition of an *invalidating* node, which models the status of the sensor as working or defective. This node provides a qualitative explanation of inconsistent data: it is caused by a defective sensor. The connection of successive instances of the invalidating node models the status of a sensor over time, allowing the DBN to handle both persistent and intermittent faults.

## 1 INTRODUCTION

A robot vehicle is to be monitored as it executes a sequence of tasks against a schedule. People and other robots cross its path, so the schedule is not strictly adhered to. On occasions the robot fails; it is late arriving at its next port of call, or it turns left instead of right. In the current application we use data from a simple sensor, a *light beam sensor*, which signals when some object crosses it. Other sensors are available also for conventional (quantitative) control and will be incorporated later into our framework for discrete probabilistic monitoring.

The conventional *quantitative* approach to such a tracking problem is to use a controller such as a Kalman Filter (Bar-Shalom and Fortmann, 1988), which is based on the cycle: predict state, measure (i.e. sense), update state estimate. Such quantitative methods are inadequate for handling the gross changes that are the focus of our work, as they are restricted to reporting ever larger covariances. Light beam sensors provide coarse, comparatively sparse data about movement in an environment, which are not suited to a conventional quantitative treatment. A symbolic representation of change is more informative, as we apply probabilistic reasoning techniques to monitoring gross changes.

Belief Networks (Pearl, 1988) integrate a mechanism for inference under uncertainty with a secure Bayesian foundation. Belief networks have been been used in various applications, such as medical diagnosis (Spiegelhalter et al., 1989) and model-based vision (Levitt et al., 1989), which initially were more static, i.e. essentially the nodes and links do not change over time. Such approaches involve determining the structure of the network; supplying the prior probabilities for root nodes and conditional probabilities for other nodes; adding or retracting evidence about nodes; repeating the inference algorithm for each change in evidence. There has also been work on the dynamic *construction* of belief networks (Breese, 1989) (Charniak and Goldman, 1989), but the desired output is still a single static network. Only recently have a few researchers used belief networks in *dynamic* domains, where the world changes and the focus is reasoning over time. Such dynamic applications include robot navigation and map learning based on *temporal* belief networks (Dean and Wellman, 1991) and monitoring diabetes (Andreassen et al., 1991). For such applications the network grows over time, as the state of each domain variable at different times is represented by a *series* of nodes. These dynamic networks are Markovian, which constrains the state space to some extent, however it is also crucial to limit the history being maintained in the network. We have developed such a dynamic belief network for discrete monitoring using light beam sensor data (Nicholson, 1992) which we briefly describe in Section 2.

Sensor data may be noisy or incorrect. In Section 3 we review how conventional quantitative methods validate sensor data and reject incorrect data, then de-scribe the types of incorrect data which may occur in the domain. In Section 4 we show how the basic DBN, which infers only an impossible combination of evidence, may be modified to handle (and implicitly reject) specific types of incorrect data. We then present an extension to the DBN in Section 5 which provides a qualitative explanation of inconsistent data being caused by a defective sensor, allowing us to model either intermittent or persistent faults.

## 2 THE DOMAIN

### 2.1 THE DISCRETE SPATIAL AND TEMPORAL MODEL

The environment (a laboratory in which a robot vehicle roams) is divided into regions by the light beam sensors. Without significant loss of generality, we restrict attention initially to rectangular regions. We monitor moving objects, which may be robots or people. An object's position is given by the region it is believed to be in. Each light beam sensor provides data about a light beam sensor crossing (BC): the direction of the crossing, and the $t_{begin}$ and $t_{end}$ time points for the time interval over which it occurred. The temporal representation is a time line divided at irregular intervals by the $t_{begin}/t_{end}$ time points. The time intervals between observation data, during which there is no change in the world state, are labelled $T_0$, $T_1$, ... $T_i$, $T_{i+1}$, etc. (We also refer to successive intervals as T and T+1.) The discrete trajectory for an object is a sequence of region/time interval pairs (R, T). The heading of an object in a given region indicates from which direction it entered (i.e. one of N, S, E, W). We also assume that we have a model of the object's *mobility*, the tendency of an object to move. This is a function over time of the speed of the object, the spatial layout, the type of object, and so on, and gives us the probability that it will move at time instant $t$, which can be projected onto the time interval T.

### 2.2 THE DBN MODEL

This discrete spatial and temporal model of the domain may be represented by the discrete valued nodes in the DBN. For now we make the reasonable practical assumption that the environment is closed and that the number of objects, $N$, in the environment, is known and fixed; the more general case is not significantly more complex but involves dynamic modification of the network structure. The spatial layout of regions and sensors is fixed and known; let us suppose that we have $M$ regions and $P$ sensors. Table 2.1 gives a summary of the types of nodes, their states, and their function in the network. The world nodes are those that represent the world state space: object position (OBJ), heading (HEAD) and mobility (MOTION), and region occupancy information (#R). Nodes are time-stamped with the time interval, $T_i$, over which they apply and during which the world

Table 1: DBN Node Types

| Node | States and Function |
|---|---|
| $OBJ_i(T)$ | $r_1, r_2, \ldots r_M$ |
| | Position of object $i$ at time T |
| $HEAD_i(T)$ | $u, h_N, h_S, h_E, h_W$ |
| | Heading of object $i$: $u$ = unknown; |
| $MOTION_i(T)$ | stat, move |
| | Mobility of object $i$; |
| | stat short for stationary |
| $\#R_j(T)$ | $0, 1, \ldots n$ |
| | Region occupancy - number only: |
| | for $i$, region $j$ contains $i$ objects |
| $BC\text{-}OBS_i$ | nc,dir1,dir2 |
| | crossing data from sensor $i$: |
| | nc indicates no crossing, |
| | dir1, dir2: the possible directions |
| $BC\text{-}ACT_i$ | nc,dir1,dir2,both |
| | Actual crossings of LB $i$; |
| | both represents objects crossing |
| | in both directions |

does not change. Observation nodes are those representing the sensor crossings: a node for the crossing data provided by the sensor (BC-OBS), and a node representing the actual physical crossing of the whole sensor which occurred (BC-ACT). We make this distinction between actual and observed because objects may cross a sensor in both directions during the observation data time interval, $T_{BC}$, however the sensors only detect a single directional crossing. The probability distribution (PD) for BC-OBS$_x$ is:

$P(BC\text{-}OBS_i = dir1 \mid BC\text{-}ACT_i = dir1) = 1$
$P(BC\text{-}OBS_i = dir2 \mid BC\text{-}ACT_i = dir2) = 1$
$P(BC\text{-}OBS_i = dir2 \mid BC\text{-}ACT_i = both) = 0.5$
$P(BC\text{-}OBS_i = dir1 \mid BC\text{-}ACT_i = both) = 0.5$
$P(BC\text{-}OBS_i = nc \mid BC\text{-}ACT_i = nc) = 1$

During any given time interval T when nothing has changed, there will be $N$ object position, heading, and mobility nodes, and $M$ region nodes. If there are $P$ light beam sensors, a sensor crossing will generate $P$ actual crossing (BC-ACT) and observed signal (BC-OBS) nodes.

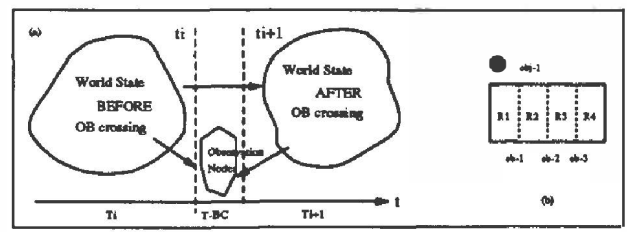

Figure 1: (a) General expansion of the network; (b) Example scenario used throughout this paper.

The dynamic construction of a network combines the world model (movement of objects between regions) and the observation model (the light beam sensor data which is generated) as the network grows over time (see Figure 1(a)). The network expansion and inference

algorithm is:

1. Make new instances of world nodes (OBJ, HEAD, MOTION, R) for the T+1 interval.
2. Connect old (T) and new (T+1) world nodes.
3. Create new observation nodes (BC-ACT, BC-OBS).
4. Connect world and observation nodes.
5. Add data as evidence for obs. nodes (BC-OBS).
6. Run inference algorithm to update beliefs.

If step 5 is omitted then the predictions made by the network corresponds to a prediction of the position of an object dependent only on its previous position and its mobility. If the sensor crossing data is added as evidence, then the inference gives an updated estimate of the object position at time interval T+1, and may also change beliefs about any node in the network, including those before time interval T.

The DBN is multiply-connected, requiring complicated inference algorithms, such as conditioning or clustering (Pearl, 1988). The problem of inference for such networks is NP-hard (Cooper, 1990), however improved algorithms such as (Jensen et al., 1990) have made inference in carefully structured networks feasible (Andreassen et al., 1987). The DBN as described gives us an inference engine which infers alternative world models (the position of object in regions) with associated probabilities, from both the model of object motion (the prior probabilities for the objects' mobility) and the the sensor crossing data (the observation model). (Nicholson, 1992) provides more details.

The example scenario, shown in Figure 1(b), used throughout this paper is a linear arrangement of 4 regions, 3 light beam sensors containing one object. The methods described in this paper also apply to multiple objects and other divisions of the environment, including a grid of sensors.

## 3  INCORRECT DATA

### 3.1  Quantitative Methods

Quantitative approaches to tracking and sensor validation involve a *noise model*; random perturbations which usually have only a small effect. For the usual case of unbiased estimators, a Gaussian model is adequate and is optimal for white noise. (If the estimators are biased, there are models for "coloured" noise). These are used for continuous variables. If variables values are discrete a Poisson distribution is used instead of a Gaussian. To handle gross errors of the sort that are the focus here, a number of different techniques have been proposed. A threshold called a *validation gate* may be applied to the Gaussian (for example, 2 standard deviations, corresponding to 97.7%). Alternatively robust statistics (Huber, 1981; Durrant-Whyte, 1987) may be used, where, for example, the error is a linear combination of two Gaussians. Finally non-parametric statistics have been developed; but they are more difficult to compute and analyse since they are non-linear.

A previous paper (Nicholson and Brady, 1992) shows how the DBN may be extended to maintain a limited history of the movement of the object. This provides a solution to the *data association problem* (DAP), that of deciding which object has given rise to which observation. Quantitative solutions to the DAP include certain techniques for handling observations which do not fall within the validation regions. One method is to discard them as "clutter", which is sometime called a *false alarm* (Bar-Shalom and Fortmann, 1988). An alternative is to initiate a new track (and hence filter) for such an observation and discontinue it after a certain time if no further data supports this hypothesis of a new object.

In some quantitative methods track continuation (Bar-Shalom and Fortmann, 1988p. 255) is done to handle missing data. If the validation region is empty, the track is extrapolated. If a predetermined number of subsequent validation regions in a row are also empty, the track is dropped.

### 3.2  Incorrect Data for the Domain

Incorrect light beam sensor data may be classified as follows:

1. *Ghost Data*: a sensor crossing is signaled but in fact never took place, a false positive. This corresponds to clutter, noise or general false alarms in quantitative methods.

2. *Wrong Direction Data*: a beam is broken, but the signaled direction of crossing is incorrect. The sensor data is inaccurate, rather than completely wrong; the sensor is certainly malfunctioning.

3. *Missing Data*: an object moves from one region to another but no sensor crossing data is registered, a false negative. This corresponds directly to missed detection in quantitative methods.

4. *Wrong Time Data*: a sensor crossing does occur, and the direction is correct, however the time stamp is incorrect.

Suppose we know that the object is in region $R_1$ at time T. The observation BC-OBS$_3$ of either direction of crossing must be ghost data. However if we know the object is in region $R_1$ and the next data received is BC-OBS$_1$ = dir2, then this may be either ghost data or wrong direction data. Obviously we are not always able to determine immediately that data is incorrect: this may depend on the *combination* of data received. Suppose that we do not know the whereabouts of the only object in the environment and that we receive two pieces of data: BC-OBS$_1$ = dir1 and BC-OBS$_3$ = dir2. Received together, the two observations are not mutually compatible, they are *inconsistent*; one must be a ghost crossing (or they both might be). If they are observed sequentially, there may have been some missing crossings, or again one or both are ghost data. We want to represent these as possible but competing alternatives, and to allow subsequent data to support a particular alternative. In this paper we do not deal with the possibility that both ghost and wrong direction data could be caused by an object which the system does not know about; we assume that all initialisation information is correct and that no new objects appear. The main point to be noted for both ghost and wrong direction data is that there is an observation node with evidence in the DBN which directly represents the incorrect data.

We have based the DBN on the assumption that the time frames are determined by the sensor data which corresponds to a change of state, i.e. an object has moved between regions. Missing data means that an object has moved undetected to another region. In some situations we can model this missing data within the existing DBN expansion and inference algorithm. Suppose, for example, there is a missing crossing for sensor $LB_1$, and an observation is received for another sensor, $LB_2$. While adding the received observation, $BC\text{-}OBS_2 = \text{dir1}$, we create a negative data node for the first sensor, $BC\text{-}OBS_1 = \text{nc}$, which represents the missing crossing (although with incorrect time stamp). However if nothing has changed, the network has not been expanded, and there is not even an incorrect nc signal recorded. If the object that made the undetected movement generates the next positive observation, then there will never be a BC-OBS added with evidence nc that actually represents the wrong reading. If the region the object has moved into undetected is otherwise unoccupied this may cause a subsequent detected sensor signal that would be considered ghost data, or wrong direction data. The higher level reasoning and additional expansion of the DBN which is required to handle this missing data is given in (Nicholson, 1992).

If the time stamp is incorrect but the temporal *order* of the observation data nodes added to the network is correct, then wrong time data will only affect the system's temporal reasoning, for example comparing against schedules and predictions. If the error in the time stamp is wrong to the extent the order of the BC nodes is wrong, this will generate problems of missing data and ghost crossings. Such incorrect ordering of data cannot be handled within the network and is not considered in this paper.

## 4 HANDLING INCORRECT DATA WITHIN THE BASIC DBN

The basic DBN does not handle inconsistent data; it finds the evidence impossible and rejects it. We can modify the existing DBN to provide a mechanism for handling certain kinds of inconsistent data.

### 4.1 MODIFYING THE PD FOR BC-OBS

The first three types of incorrect data which we identified above involve a discrepancy between the sensor crossing data received by the DBN controller, and the crossing which actually took place. We have already modeled the distinction between the crossing which took place and the data received by creating the two types of sensor crossing node, BC-ACT and BC-OBS. The modification to the existing DBN involves changing the probability distribution for the BC-OBS node. Instead of using binary values, we represent the uncertainty in the network itself, as the PD entries for each $BC\text{-}OBS_x$ become:

$P(BC\text{-}OBS=\text{dir1}|BC\text{-}ACT=\text{dir1})=conf_1$    ok
$P(BC\text{-}OBS=\text{dir2}|BC\text{-}ACT=\text{dir1})=(1-conf_1)/2$    wrong
$P(BC\text{-}OBS=\text{nc}|BC\text{-}ACT=\text{dir1})=(1-conf_1)/2$    miss.

$P(BC\text{-}OBS=\text{dir1}|BC\text{-}ACT=\text{dir2})=(1-conf_1)/2$    wrong
$P(BC\text{-}OBS=\text{dir2}|BC\text{-}ACT=\text{dir2})=conf_1$    ok
$P(BC\text{-}OBS=\text{nc}|BC\text{-}ACT=\text{dir2})=(1-conf_1)/2$    miss.

$P(BC\text{-}OBS=\text{dir1}|BC\text{-}ACT=\text{nc})=(1-conf_2)/2$    ghost
$P(BC\text{-}OBS=\text{dir2}|BC\text{-}ACT=\text{nc})=(1-conf_2)/2$    ghost
$P(BC\text{-}OBS=\text{nc}|BC\text{-}ACT=\text{nc})=conf_2$    ok

The confidence in the observation is given by some value based on a model of the sensor's performance and is empirically obtainable; $conf_1$ is the confidence in the positive sensor data, $conf_2$ is the confidence in the negative sensor data (or, $1\text{-}conf_2$ is the probability of ghost data). We have modeled positive data being ghost or wrong direction data as being equiprobable - this need not be the case and can be replaced by any alternative plausible values. Likewise for negative data, although the equiprobable direction of the actual crossing seems intuitively reasonable.

### 4.2 RESULTS FOR UNINITIALISED EXAMPLE

We now show the results from the modified DBN for the example environment, with the position of the object at $T_0$ unknown. The sensor observations made are as follows.

| Crossing | $BC\text{-}OBS_1$ | $BC\text{-}OBS_2$ | $BC\text{-}OBS_3$ |
|---|---|---|---|
| $T_0$ to $T_1$ | dir1 | nc | nc |
| $T_1$ to $T_2$ | nc | nc | dir2 |
| $T_2$ to $T_3$ | nc | nc | dir1 |
| $T_3$ to $T_4$ | nc | nc | dir2 |

Table 2 and Figure 2 shows the beliefs inferred by the DBN after each new observation is received and the network expanded. Each row of example diagrams shows the updated beliefs for the position of the object at some time T. The observations are shown between the appropriate rows. Each column corresponds to the belief at some time T for the position of the object over time, i.e. shows the inferred trajectory. We make the following observations on these results.

Table 2: Beliefs inferred by the modified DBN for inconsistent observations. Initial position unknown. $conf = 0.99$.

| Node | State | $T_1$ | $T_2$ | $T_3$ | $T_4$ |
|---|---|---|---|---|---|
| $OBJ_1(0)$ | $R_1$ | 0.8827 | 0.4781 | 0.0766 | 0.0042 |
|  | $R_2$ | 0.0405 | 0.02195 | 0.0035 | 0.0002 |
|  | $R_3$ | 0.0384 | 0.0221 | 0.0403 | 0.0436 |
|  | $R_4$ | 0.0384 | 0.4779 | 0.8796 | **0.9520** |
| $OBJ_1(1)$ | $R_1$ | 0.0405 | 0.0219 | 0.0035 | 0.0002 |
|  | $R_2$ | 0.8827 | 0.4781 | 0.0766 | 0.0042 |
|  | $R_3$ | 0.0384 | 0.0220 | 0.0401 | 0.0433 |
|  | $R_4$ | 0.0384 | 0.4780 | 0.8798 | 0.9523 |
| $OBJ_1(2)$ | $R_1$ |  | 0.0221 | 0.0035 | 0.0002 |
|  | $R_2$ |  | 0.4779 | 0.0764 | 0.0039 |
|  | $R_3$ |  | 0.4781 | 0.9162 | 0.9918 |
|  | $R_4$ |  | 0.0220 | 0.0039 | 0.0041 |
| $OBJ_1(3)$ | $R_1$ |  |  | 0.0035 | 0.002 |
|  | $R_2$ |  |  | 0.0763 | 0.0039 |
|  | $R_3$ |  |  | 0.0768 | 0.0041 |
|  | $R_4$ |  |  | 0.8434 | 0.9918 |
| $OBJ_1(4)$ | $R_1$ |  |  |  | 0.0002 |
|  | $R_2$ |  |  |  | 0.0039 |
|  | $R_3$ |  |  |  | 0.9526 |
|  | $R_4$ |  |  |  | 0.0433 |

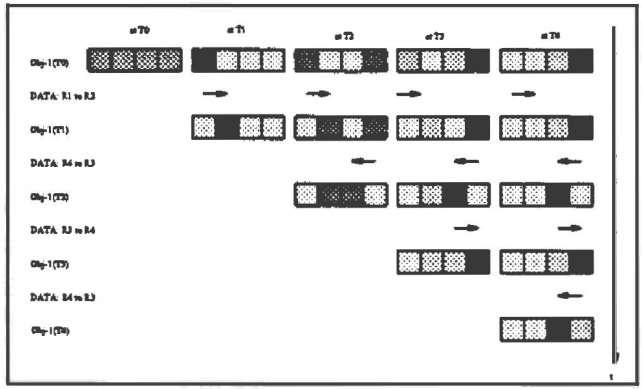

Figure 2: Object position beliefs for inconsistent observations from Table 2. The belief that the object is in a region is indicated by the intensity of shading.

**Beliefs during $T_0$:** 4 alternatives are being maintained explicitly, all equally probable.

**Beliefs during $T_1$:** The DBN is nearly certain that the OBJ moved $R_1$ to $R_2$. The initial beliefs (i.e. the 0th instance) have been revised, indicating that the OBJ was very likely to have been in $R_1$. If the data was ghost data (considered unlikely), there is a small chance that the object started in $R_2$, $R_3$ or $R_4$. There is also the alternative that the crossing occurred but in the opposite direction. Hence the belief for $R_2$ (ghost plus wrong direction alternatives) is larger than $R_3$ and $R_4$ (ghost only).

**Beliefs during $T_2$:** The system now maintains the alternatives:

| $BC\text{-}OBS_1(T_1)$ | $BC\text{-}OBS_3(T_2)$ |
|---|---|
| correct | ghost |
| ghost | correct |
| wrong direction | ghost |
| ghost | wrong direction |
| ghost | ghost |

The first two alternatives have the same probabilities and are considered the most likely (i.e. approaching 0.5 probability).

**Beliefs during $T_3$:** The additional $BC\text{-}OBS_3(T_3)$ crossing (from $R_3$ to $R_4$) acts as support for the $BC\text{-}OBS_3(T_2)$ observation being correct; **belief**($BC\text{-}ACT_3(T_2)$ = dir2) = 0.8761 and **belief**($BC\text{-}ACT_1(T_1)$ = nc) = 0.9265, i.e. $BC\text{-}OBS_1(T_0)$ was probably ghost data.

**Beliefs during $T_4$:** The additional $BC\text{-}OBS_3(T_4)$ crossing is further support for the alternative that the first observation was ghost data and second correct; **belief**($BC\text{-}ACT_1(T_1)$=nc) = 0.996 and **belief**($BC\text{-}ACT_3(T_2)$=nc) = 0.9484. The DBN has inferred that the object is probably initially in $R_4$; **belief**($OBJ_1(T_0)$=$r_4$) = 0.9520.

### 4.3  RESULTS FOR INITIALISED EXAMPLE

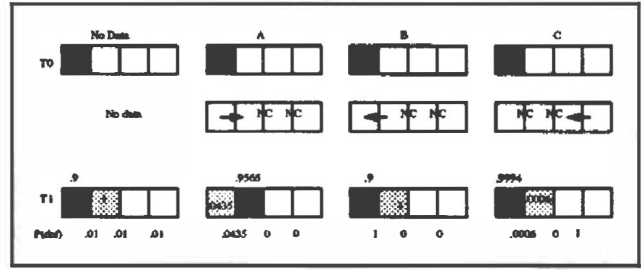

Figure 3: Object position beliefs with successive BC-INV nodes unconnected.

Figure 3 shows the beliefs inferred with the object initially in $R_1$, with no observations added (column 1), then for the 3 alternative observations shown. We make the following observations on these results.

**No observations:** The object may stay in $R_1$ or move into $R_2$. Sensors $LB_2$ and $LB_3$ should generate a no-crossing signal, because there are no object in the regions they separate, however the DBN infers a small probability of a ghost crossing signal. The beliefs inferred for the signal $BC\text{-}OBS_1$ are a combination of the possible nc or dir2, plus possible incorrect data from the sensor, hence the predicted observation probabilities differ from the actual crossings predicted.

**Observation A:** For the $BC\text{-}OBS_1$ = dir1 crossing data, the DBN correctly infers that this might be correct data (i.e. $BC\text{-}ACT_1$ = dir1) or ghost data (i.e. $BC\text{-}ACT_1$ = nc).

**Observation B:** Because there was no object in $R_2$ at $T_0$, the BC-ACT$_1$ = dir2 crossing must be incorrect data; it may be either ghost data, or wrong direction data.

**Observation C:** The DBN infers that BC-ACT$_3$ must be nc, implicitly rejecting the observation as incorrect data.

### 4.4 Using Virtual Evidence

Our model includes observations as specific evidence for a variable, the BC-OBS node. One possible alternative would be to model the uncertainty in the accuracy of the observation by using *virtual evidence* (Pearl, 1988), which is given as a likelihood ratio of the states of the BC-OBS node. If the data was for a dir1 crossing of sensor LB$_x$, then the specific evidence using the existing scheme would be set-evidence(BC-OBS$_x$ = dir1). The corresponding virtual evidence for takes the form dir1:DIR2:NC, i.e. $conf:(1-conf)/2:(1-conf)/2$. This use of virtual evidence provides the same results as modifying the PD for BC-OBS. Since the BC-OBS evidence is the physical output of a sensor, we prefer to enter it as specific evidence and model the difference between the observation from the sensor and the actual crossing within the DBN itself.

## 5 EXPLAINING BAD DATA AS A DEFECTIVE SENSOR

The modification to the DBN described in the previous section provides a mechanism for handling (by implicitly rejecting) certain inconsistent data. It represents adequately the underlying assumptions about the data uncertainty, which are that the observed sensor crossing might not match the actual sensor crossing that took place. However it does not provide an explanation of *why* the observed sensor data might not reflect the actual crossing. We want to represent the most usual source of incorrect data, namely a defective sensor.

### 5.1 THE INVALIDATING NODE

We adapt an idea that has been used in other research areas, that of a moderating or invalidating condition. In the social sciences and psychology, the term "moderator" is used for an alternative variable that "messes up" or "moderates" the relationship between other variables (Zedeck, 1971; Wermuth, 1987; Wermuth, 1989). A similar idea has been used in expert system research; in (Andersen et al., 1989) such nodes are called "invalidators". Of course, this idea is also familiar to the AI community; Winston (Winston, 1977) described the notion of a *censor*, which acts as an "unless" condition: *if* a BC-ACT occurs, *then* BC-OBS will be generated *unless* the sensor is defective.

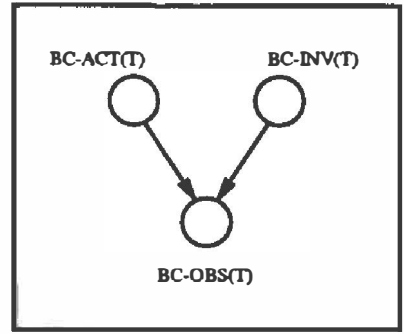

Figure 4: Adding the invalidating node, BC-INV, to the DBN

We add a node, BC-INV, the *invalidating* node, which has two states, [work, def], short for "working" and "defective". It is connected as a predecessor of BC-OBS (see Figure 4). The PD for BC-OBS for ghost data, wrong direction data, and missing data is given by:

P(BC-OBS=dir1 | BC-ACT=dir1 BC-INV=work) = 1
P(BC-OBS=dir2 | BC-ACT=dir2 BC-INV=work) = 1
P(BC-OBS=nc | BC-ACT=nc BC-INV=work) = 1

P(BC-OBS=dir2 | BC-ACT=dir1 BC-INV=def) = 0.5
P(BC-OBS=nc | BC-ACT=dir1 BC-INV=def) = 0.5
P(BC-OBS=dir1 | BC-ACT=dir2 BC-INV=def) = 0.5
P(BC-OBS=nc | BC-ACT=dir2 BC-INV=def) = 0.5
P(BC-OBS=dir1 | BC-ACT=nc BC-INV=def) = 0.5
P(BC-OBS=dir2 | BC-ACT=nc BC-INV=def) = 0.5

The question then arises: what are the prior probabilities for BC-INV? We explicitly represent how likely it is that the sensor is working correctly by the prior probabilities for BC-INV, which can be obtained from empirical data; $conf$ is now explicitly the confidence that the sensor is working.

P(BC-INV=work | ) = $conf$
P(BC-INV=def | ) = 1-$conf$

### 5.2 RESULTS FOR SENSOR STATUS

The inference algorithm was run for the same set of alternative observations (A, B and C) on the DBN with the BC-INV nodes added; again the object is initially in $R_1$ and $conf = 0.99$. The additional beliefs inferred for the BC-INV nodes having state def are shown in Figure 3 under the appropriate sensor (in row labelled P(def)). For cases A and B, the DBN infers that all nc observations for sensors $LB_2$ and $LB_3$ are correct (i.e. BC-INV = work) because there were no objects in adjacent regions to move across these sensors. In case C, the DBN infers correctly that sensor $LB_3$ must be defective (i.e. BC-INV$_3$ = def); there is a small possibility that the nc observation for sensor $LB_1$ may be incorrect, if there is missing data.

### 5.3 MODELING SENSOR STATUS OVER TIME

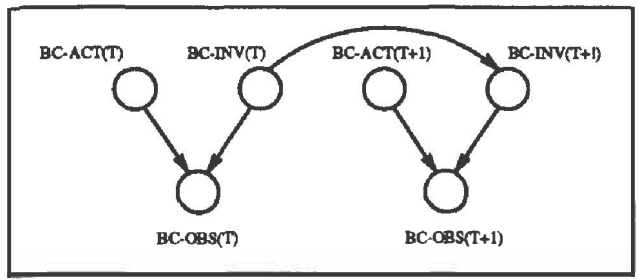

Figure 5: Modeling sensor status over time.

The invalidating node provides the explicit representation of the cause of incorrect data - a defective sensor. However, there is no connection between successive BC-INV nodes, which means no correlation between the working status of a sensor at different times. If the DBN infers that a sensor is defective at some time T because the data received has been wrong, it should also effect the interpretation put on subsequent (and possibly earlier) data from that sensor. To provide such a model of the sensor, we assume that at the initial time $T_0$ all BC-INV$_i$($T_0$) nodes have some prior such as described above. At each time step, a copy is made of all the BC-INV nodes (whether or not any data is received for that sensor), and each is connected to its successor (see Figure 5). The PD for each BC-INV(T+1) is then given by:

P(BC-INV(T+1)=work | BC-INV(T) = work) = 1 - $d$
P(BC-INV(T+1)=def  | BC-INV(T) = work) = $d$
P(BC-INV(T+1)=def  | BC-INV(T) = def)  = 1-$X$
P(BC-INV(T+1)=work | BC-INV(T) = def)  = $X$

where $d$ is a degradation factor and $X$ is related to the consistency of the fault.

The degradation factor $d$ is the probability that a sensor which has been working during the previous time interval has begun to respond defectively. It is based on a model of the expected degradation of the sensor and is a function of the time between sensor readings.

### 5.4 PERSISTENT AND INTERMITTENT FAULTS

There are two general models for a defective sensor: an intermittent fault, which means that not every signal from the sensor is incorrect; a persistent fault, that manifests itself for each observation.

One method for modeling an intermittent fault is to make the variable $X$ strictly positive. However if the DBN infers from the data that (i) BC-INV($T_i$) = def, and (ii) BC-INV($T_{i+1}$) = work then the fault detected during $T_i$ cannot be passed on to $T_{i+2}$. An alternative is to have $X = 0$ all the time (i.e. once a sensor is known to be defective it remains defective) and change the PD for BC-OBS so that if a defective sensor can still produce correct data:

P(BC-OBS=dir1|BC-ACT=dir1 BC-INV=def)=$x > 0$
P(BC-OBS=dir2|BC-ACT=dir2 BC-INV=def)=$x > 0$
P(BC-OBS=nc|BC-ACT=nc BC-INV=def)=$x > 0$

A persistent fault may be modeled by $X$ equals 0, but without the need to change the probability distribution for BC-OBS. An example of a persistent fault is the incorrect wiring of the sensor so that the crossing direction is wrong each signal. In practice, a controller will request confirmation of the status of the sensor, or receive information that it has been repaired. In this case, BC-INV(time-of-report) will have no predecessors and the prior will reflect the confidence in the status report. Results from the DBN with the BC-INV(T) to BC-INV(T+1) connection for the same set of observations are shown in Figure 6, $conf = 0.99$, $d = 0.01$, $X = 0$. Since $X = 0$, once sensors LB$_1$ (case B) and LB$_3$ (case C) has been identified as definitely defective, the DBN infers that the probability it was defective initially (BC-INV$_3$($T_0$) = def) is 0.5025.

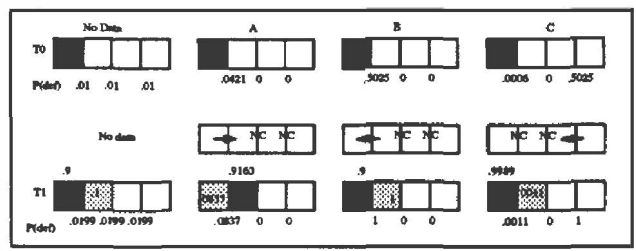

Figure 6: Object position beliefs inferred by DBN with successive invalidating nodes connected.

### 5.5 MODELING DIFFERENT DEFECTS

The current BC-INV node, with only two states, does not allow the explanation to distinguish between types of defects. We can increase the BC-INV states to [work, def-ghost, def-dir, def-miss], for ghost data, wrong direction data and missing data respectively. Details of and results for these additional states, as well as results for various combinations of $conf$, $d$ and $X$ may be found in (Nicholson, 1992).

## 6   CONCLUSIONS

The basic DBN provides discrete tracking of objects based on light beam sensor data, in a method which is analogous to quantitative filter techniques. In this paper we have described a solution to the problem of incorrect or noisy data. By changing the PD for the BC-OBS node to contain values other than 1 or 0, the DBN is able to handle inconsistent data, rather than simply inferring a contradiction in the evidence. The addition of a node which models the status of the sensor as working or defective, as another parent of the BC-OBS node, provides an explanation of the incorrect data as being caused by a defective sensor.

The connection of the successive instances of the invalidating node models the status of a sensor over time, allowing the DBN to handle both persistent and intermittent faults. We have shown that a combination of AI techniques – discrete representation and reasoning with uncertainty – can provide a solution to a real world problem, i.e incorrect sensor data. Moreover, the solution is in some ways more intuitive than equivalent conventional quantitative methods.

### Acknowledgements

We wish to thank Hugh Durrant-Whyte, Finn Jensen, Uffe Kjaerulff, Steffen Lauritzen, Stuart Russell and David Spiegelhalter for valuable discussions during the development of this research; the Rhodes Trust.